\documentclass{article}

\usepackage[preprint]{corl_2026} 
\usepackage{booktabs}
\usepackage{graphicx}
\usepackage{array}
\usepackage{amsmath}
\usepackage{amssymb}
\usepackage{placeins}
\usepackage{microtype}

\title{Nudge Before You Push: Physics-Aware Navigation via Tactile Probing}

\author{%
  Xianyao Li$^{1}$ \quad Fang Xu$^{1}$ \quad Ruitong Tian$^{1}$ \quad Bowen Sun$^{1}$ \quad Xiao Hu$^{2}$ \quad Yang Ye$^{2}$ \quad Jing Du$^{1}$ \\[2pt]
  $^{1}$Department of Civil and Coastal Engineering, University of Florida \\
  $^{2}$Department of Civil and Environmental Engineering, Northeastern University
}

\AtBeginDocument{\hypersetup{
  pdftitle={Nudge Before You Push: Physics-Aware Navigation via Tactile Probing},
  pdfauthor={Xianyao Li, Fang Xu, Ruitong Tian, Bowen Sun, Xiao Hu, Yang Ye, Jing Du},
  pdfsubject={Tactile probing for robotic navigation}
}}

\begin{document}
\raggedbottom
\maketitle


\begin{abstract}
Visually identical containers can conceal loads that require different handling decisions. We present \textsc{TANav}, which uses a brief nudge to measure push resistance for navigation under a site-defined handling boundary. \textsc{TacPhys} reads the force sequence, with optional RGB-D and kinematics, into a mass estimate for push authorization. A repeated-patrol planner weighs probe and route costs, requests a second contact when useful, and reuses observations across visits. In simulation, \textsc{TacPhys} approaches a resistance-only Bayes reference and reduces missed pushes from 28.7\% to 5.5\% relative to peak-force thresholding at comparable low-risk operating points. In repeated-patrol simulation, \textsc{TANav} recovers 90\% of the oracle's path saving, more than halves human interventions relative to always-detour, and reduces boundary violations from 4.3\% to 2.9\% relative to RGB-D-only probing. On a quadruped manipulator with a Hall-array fingertip, offline zero-shot MAE is 0.28--1.07\,kg on containers up to 2.82\,kg. Force-rise calibration at 3\,kg gives 93.5\% pooled offline accuracy (86.4\% on non-cube episodes); a separate raw-peak rule gives 15 of 20 correct online decisions on unseen boxes.
\end{abstract}


\suppressfloats[t]

\begin{figure*}[t]
    \centering
    \includegraphics[width=\textwidth]{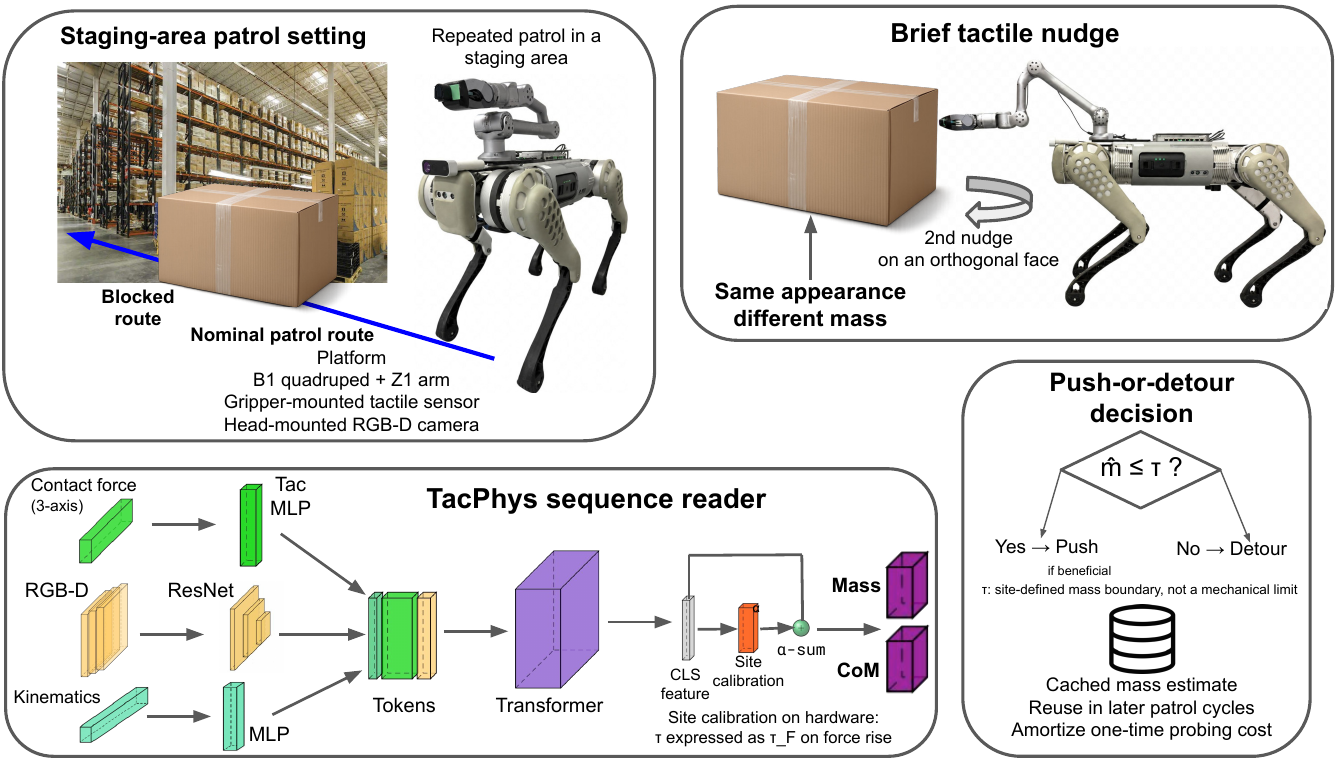}
    \caption{\textbf{Tactile probing for push-or-detour navigation.} \textsc{TANav} separates contact interpretation, push eligibility, and route choice. In simulation, \textsc{TacPhys} reads nudge resistance into mass; the site boundary determines eligibility, and the patrol planner weighs probe and route costs, requests additional contact, and reuses cached observations. Hardware tests assess offline transfer and calibration; calibrated raw-peak decisions run online.}
    \label{fig:overview}
\end{figure*}

\section{Introduction}
\label{sec:intro}

Industrial patrol robots revisit routes that may change between inspection cycles~\citep{industrial_legged_patrol,anymal_industrial_inspection}. We consider staging areas where temporarily placed cartons or totes can obstruct familiar paths. Box-like containers can conceal different payloads behind similar exteriors, while accessible, approximately planar sides allow repeatable probing from complementary directions. Before moving an unfamiliar box, a person nudges it to feel resistance. \textsc{TANav} turns this intuition into a mass estimate that informs the route-clearing decision.

The decision also differs from mechanical movability. Navigation among movable obstacles (NAMO) uses movability or manipulation cost inferred from attempts or appearance~\citep{namo_stilman_kuffner,namo_decision_theoretic}. A site instead permits clearing light containers and reserves heavier loads for operators, even if both are movable. The decision thus compares mass with a site-defined boundary auditable against known loads.

\textsc{TANav} separates contact interpretation, push authorization, and route choice (Fig.~\ref{fig:overview}). A gripper-mounted fingertip records nudge resistance; \textsc{TacPhys} reads force sequences into mass for comparison with the site boundary. The planner checks eligibility before comparing push and detour costs, requests useful additional contacts, and reuses observations across visits to amortize probing. Hardware tests assess offline mass transfer and calibrated force decisions; patrol evaluation is simulated.

The mass estimate also parameterizes simulated patrol cost. Auxiliary center-of-mass (CoM) classes describe load distribution and support analysis of the information provided by the probe.

The nudge supplies evidence for authorization. The tested try-push and appearance-prior policies reach similar autonomy but authorize 55.5\% of heavy containers in simulated patrol. RGB-D observations during the nudge leave 4.3\% violations, force sequences 2.9\%, and two-contact agreement 1.0\%. Caching observations amortizes contact cost across visits. Force sequences improve decision trade-offs over scalar summaries, with similar gains across encoders. Offline hardware tests show zero-shot transfer on light containers and deferral through contact agreement; online decisions use site calibration.

Our contributions are:
\begin{enumerate}
    \item A formulation of load-dependent navigation with a diagnostic probe, an explicit authorization boundary in mass units, and \textsc{TacPhys}, a learned sequence reader that estimates mass from push resistance.
    \item A repeated-patrol decision scheme that weighs contact and route costs, selectively acquires a second probe, and reuses observations across visits.
    \item A simulation-to-hardware evaluation across two tactile front ends, covering a resistance-only reference, offline mass transfer, site-calibrated online decisions, and simulated patrol efficiency and authorization errors.
    \item An identifiability analysis of the probe: an orthogonal second contact improves both mass and CoM estimates, and contact onset is a key protocol variable for CoM transfer to hardware.
\end{enumerate}

\section{Related Work}

\paragraph{Navigation among movable obstacles.}
NAMO extends geometric planning by allowing a robot to rearrange obstacles~\citep{namo_stilman_kuffner}, including under incomplete knowledge of movability~\citep{namo_unknown_env,namo_decision_theoretic}. Subsequent methods infer affordances, manipulation cost, or movability from observations~\citep{affordance_namo,learning_namo_cost,movability_aware_navigation}. Mobile manipulators also combine navigation with learned pushing control and interaction-state estimation~\citep{legged_pushing_policy,privileged_push_estimator}; tactile skins support negotiation of obstacles during contact~\citep{tactile_property_estimation}. We study a diagnostic contact before route clearing, using its response to evaluate an authorization boundary and comparing readers under the same planner in simulation.

\paragraph{Physical inference through contact.}
Interactive perception uses actions to reveal properties that passive sensing cannot resolve~\citep{interactive_perception}. Inertial estimation can use manipulator excitation~\citep{object_mass_estimation}, tactile in-hand interaction~\citep{grasp_and_shake_physics}, or non-prehensile pushing~\citep{nonprehensile_property_estimation,mavrakis_com_estimation}. Pushing mechanics and experimental datasets show how friction and contact geometry affect the response~\citep{mason_pushing,lynch_mason_stable_pushing,mit_push_dataset}. Optical tactile sensors provide deformation measurements~\citep{digit_sensor,gelsight}, with calibrated force regressors supporting push-based weighing~\citep{feelanyforce}; magnetic arrays provide local loading measurements~\citep{uskin,reskin}. We use a force sequence as the simulation--hardware interface and assess the information needed for a navigation decision.

\section{Method}
\label{sec:method}

\subsection{Decision and Probe}

A box-like container with an accessible side face and unknown mass \(m\) blocks the patrol route. We consider containers for which the site permits the prescribed diagnostic nudge. A site-defined handling boundary \(\tau\) governs the subsequent route-clearing push: objects with \(m\le\tau\) are eligible, while heavier objects must be bypassed or referred to an operator. We evaluate 3\,kg as the main boundary. A reader supplies an eligibility judgment, with optional deferral. For a learned mass estimate \(\hat m\),
\begin{equation}
    \pi(\hat m;\tau)=
    \begin{cases}
        \mathrm{push}, & \hat m\le\tau,\\
        \mathrm{detour}, & \hat m>\tau
    \end{cases}.
    \label{eq:decision}
\end{equation}
A deployment applies the same boundary through readings calibrated on its own fingertip (Sec.~\ref{sec:calibration}); the planner is unchanged. In simulation, the planner also checks whether an eligible clearing push reduces route cost. The robot's physical push limit separately determines whether an attempted push succeeds; we vary it independently of \(\tau\) in the patrol evaluation.

The probe precedes the route-clearing push. On hardware, the arm approaches at 3\,cm/s until contact, rests for 0.3\,s, advances 4--7\,cm at a speed sampled log-uniformly from 0.10--0.25\,m/s, dwells for 0.3\,s, and retracts. Contact position varies; median duration, including approach and retraction, is 2.6\,s. The controller stops commanded motion near joint limits; aborted probes permit operator-commanded retraction and restoration of joystick control. A second nudge can contact an orthogonal face for another view of the response. In simulation, the robot repositions to the adjacent face.

The main simulation protocol begins contact at a sampled speed of 0.15--0.35\,m/s. Additional datasets use a hardware-like creep, rest, and speed ramp to test dependence on contact onset. Same-face repetitions control for the benefit of changing geometry.

\subsection{Learned Sequence Reader: TacPhys}
\label{sec:model}

Nudge \(k\) provides force \(F^k\), kinematics \(q^k\), and an RGB-D clip \(I^k\). We align the streams to push onset on a 60\,Hz grid, retaining eight pre-onset and 96 post-onset samples; missing samples are padded and masked. Force has three components. The 37-dimensional kinematic vector contains end-effector pose and velocity, joint positions and velocities, motor torques, and position targets. Eight RGB-D frames span the push, each resized to \(224\times224\).

MLPs tokenize force and kinematics; a ResNet-style encoder projects RGB-D frames to the same width of 192. Time, modality, and nudge embeddings identify each token, and a rotation embedding describes the second contact. A six-layer, six-head transformer~\citep{attention} processes the combined sequence. Its \texttt{[CLS]} feature feeds a log-mass head and three auxiliary center-of-mass (CoM) classifiers. Each CoM class denotes the low, middle, or high third of an object axis in the first-contact frame. Only mass enters the navigation decision. The temporal encoder has 2.67M parameters; the full model has 14.2M.

For standardized log-mass \(y=(\log_{10}m-\mu_y)/\sigma_y\), with statistics fixed from training data, the loss is
\begin{equation}
    \mathcal L=(\hat y-y)^2+
    \frac{1}{3}\sum_{a\in\{x,y,z\}}
    \mathrm{CE}(\hat c_a,c_a).
\end{equation}
Training uses AdamW with learning rate \(3\times10^{-4}\), weight decay \(10^{-3}\), batch size 32, dropout 0.2, one warm-up epoch, and cosine decay over 20 epochs. Early stopping uses validation loss with patience four. We average five seeds unless specified otherwise. Masking streams gives tactile-only, visual-only, and fused variants.

\subsection{Site Calibration and Agreement}
\label{sec:calibration}

Two readings implement the site boundary. \textsc{TacPhys} supplies continuous mass estimates for simulated patrol cost and reuse when \(\tau\) changes. A one-parameter force-rise boundary fitted on known loads is auditable with a scale and supports authorization beyond the fingertip's mass-resolution range, near 4\,kg here. An affine map corrects learned mass on the same loads; both calibrations are fitted once per site and fingertip. Both domains provide force in newtons. We evaluate an optical DIGIT fingertip with a learned force regressor~\citep{feelanyforce} and a 120-taxel Hall array (MagPad), resampled from 36\,Hz to the model's 60\,Hz grid. The array sum is scale-calibrated at 20.7 counts/N (\(R^2=0.94\)). Hall-array mass evaluations subtract each nudge's mean force vector over valid samples in the eight-sample pre-onset window. Affine calibration fits \(\log_{10}\hat m'=a\log_{10}\hat m+b\) on calibration instances.

The force boundary uses the increase in normal force during the push:
\begin{equation}
    s_k=\max_{t\in\mathcal W_{\rm push}}F_n^k(t)
    -\frac{1}{|\mathcal W_{\rm pre}|}
       \sum_{t\in\mathcal W_{\rm pre}}F_n^k(t),
    \label{eq:rise}
\end{equation}
where \(\mathcal W_{\rm push}\) is the first second after push onset and \(\mathcal W_{\rm pre}\) contains the preceding contact samples. Subtracting this preload reduces variation from the approach. The boundary \(\tau_F\), fitted to mass-boundary labels on known loads, classifies the two-contact mean \(\bar s\).

Agreement accepts only matching contact judgments:
\begin{equation}
    \pi_{\rm agree}(s_1,s_2)=
    \begin{cases}
        \mathrm{push}, & \max(s_1,s_2)<\tau_F,\\
        \mathrm{detour}, & \min(s_1,s_2)\ge \tau_F,\\
        \mathrm{defer}, & \text{otherwise}.
    \end{cases}
    \label{eq:agreement}
\end{equation}
For learned per-contact agreement, the single-nudge model reads each contact separately. Matching judgments permit use of the joint two-nudge estimate; disagreement triggers deferral. Peak-force agreement uses separate peak-based estimates with the same acceptance rule. Sequential agreement compares the learned first-nudge and joint estimates.

\subsection{Repeated-Patrol Planner}
\label{sec:planner}

The simulation reader comparison uses a common NAMO-style planner~\citep{namo_decision_theoretic}. It pushes only if the object is eligible, the geometry permits clearance, and expected push cost is below detour cost. Let \(C_P\) be the cost of a successful clearing push, \(C_D\) the cost of detouring or requesting assistance, and \(C_F\) the setup and failure cost before fallback. Given an estimated probability \(p_H\) that the object exceeds the handling boundary, the planner compares
\begin{equation}
    \bar C_P=(1-p_H)C_P+p_H(C_F+C_D)<C_D.
    \label{eq:push_cost}
\end{equation}
The main study sets the physical push limit to \(\tau\), so violating pushes fail. When that limit is raised, violations and failures separate; the cost model still prices a boundary violation as failure, a conservative approximation.
Its probability model uses a lognormal error scale \(\sigma\) for each reader, estimated on validation predictions and frozen before patrol evaluation:
\begin{equation}
    p_H=1-\Phi\!\left(\frac{\log\tau-\log\hat m}{\sigma}\right),
\end{equation}
where \(\Phi\) is the standard normal cumulative distribution. The simulated observations are held-out model predictions. The empirical error model supports cost comparison; authorization still requires Eq.~\eqref{eq:decision}.

Before contacting a new blocker, the planner uses the prior probability \(p_L\) of an eligible load to test whether probing is worthwhile:
\begin{equation}
    p_L(C_D-C_P)R>C_N,
    \label{eq:probe_value}
\end{equation}
where \(C_N\) is probe cost and \(R=\min(K,1/p_{\rm rep})\) approximates reuse over the \(K\) remaining patrols when replacement probability is \(p_{\rm rep}>0\). When no replacement occurs, \(R=K\). This rule spends contact time where finding a light object can save travel or assistance on subsequent visits.

A deferred judgment leads to a detour, or to human assistance when no detour exists. Estimates are cached with the obstacle's map entry. Correct object reassociation and cache invalidation on replacement are supplied by the simulator as assumptions. An adaptive variant requests a second nudge when \(|\log\hat m-\log\tau|<\sigma\). The planner therefore treats an additional contact as a decision cost whose benefit depends on the first estimate and the route.

\section{Experiments}
\label{sec:experiments}

\subsection{Data and Evaluation}

\textbf{Simulation.} We collect 51{,}200 paired episodes in Isaac Lab~\citep{isaac_lab} using a B1+Z1 digital twin and procedural cuboids; 48{,}731 pass checks for complete contacts, valid observations, and stable resets. Mass is sampled log-uniformly over \([0.5,15]\)\,kg, independently of size, appearance, and object and floor friction. Edge ranges are 0.3--0.5, 0.3--0.6, and 0.3--0.9\,m. Static and dynamic friction ranges are 0.3--0.9 and 0.2--0.8, with dynamic friction capped by static friction and arithmetic averaging at contact. Contact location, speed, CoM, and lighting vary. Second contacts are orthogonal with probability 0.75 and otherwise use the same face. Train/validation/test splits are 80/10/10 by collection session, with both contacts kept together.

\begin{figure}[t]
    \centering
    \includegraphics[width=\linewidth]{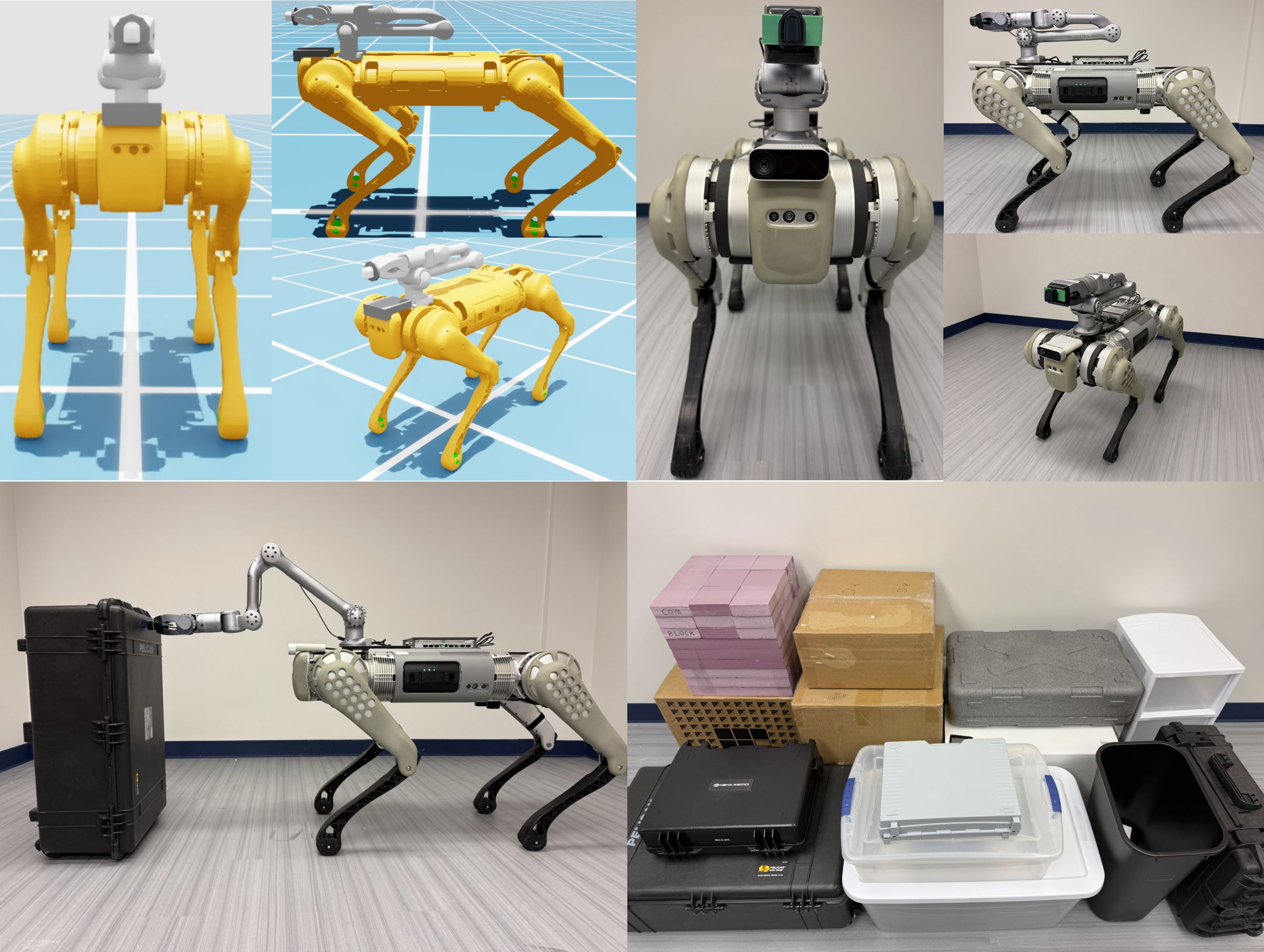}
    \caption{Platform and collection setup. Top: simulated and real B1+Z1 with the fingertip and head-mounted RGB-D camera. Bottom: a hardware contact and representative objects from the real-world dataset.}
    \label{fig:platform}
\end{figure}

\textbf{Hardware.} A Unitree B1 with a Z1 arm carries the fingertip and an Azure Kinect (Fig.~\ref{fig:platform}). The Hall-array set contains 262 clean paired episodes: 196 on a 1.69\,kg cube with 15 CoM configurations, and 66 on 11 mass instances of six containers spanning 1.76--10.52\,kg. These are five cardboard boxes at two loads each and a square plastic waste container, recorded on carpet and a plastic sheet. Of the 262 pairs, 220 use an orthogonal second face and 42 repeat on the same face as controls. Archived zero-shot and affine mass estimates use the 220 orthogonal pairs, matching the evaluated model's training protocol. The calibrated boundary and agreement rules apply to both protocols and use all 262 pairs. The optical set contains 554 single nudges on 13 objects spanning 0.74--12.96\,kg and a calibration cube, with 480 retained.

Optical single nudges support the force-readout comparison (Fig.~\ref{fig:sensors}); Hall-array data also support calibration and online evaluation. Each front end's force--mass response is characterized on its own collection.

\textbf{Metrics and splits.} We report mass MAE and accuracy at \(\tau=3\)\,kg. Decision risk is the fraction of heavy objects authorized to be pushed; missed pushes are the fraction of eligible objects classified for detour. Selective accuracy and risk use accepted episodes, with coverage reported separately. Simulation decision comparisons share 3{,}618 orthogonal test episodes. Mass metrics use each configuration's eligible test cohort. Hardware folds hold out mass instances or containers; instance folds may retain another load of the same container.

\subsection{Estimation and Decision Quality in Simulation}
\label{sec:sim}

\begin{table}[t]
\centering
\caption{Simulation estimation and decisions. MAE is mean \(\pm\) standard deviation over five seeds; decision accuracy uses the common orthogonal test cohort.}
\label{tab:estimation}
\footnotesize\setlength{\tabcolsep}{3pt}
\begin{tabular}{lccc}
\toprule
Input & Nudges & MAE (kg) & Acc. (\%) \\
\midrule
RGB-D & 1 & \(0.963\pm0.019\) & 92.2 \\
Tactile & 1 & \(0.757\pm0.007\) & 94.5 \\
Tactile + RGB-D & 1 & \(0.729\pm0.006\) & 94.7 \\
RGB-D & 2 & \(0.887\pm0.037\) & 92.9 \\
Tactile & 2 & \(0.665\pm0.008\) & 95.3 \\
Tactile + RGB-D & 2 & \(\mathbf{0.652\pm0.006}\) & 95.4 \\
Tactile + RGB-D + kin. & 2 & \(0.660\pm0.004\) & 95.5 \\
\bottomrule
\end{tabular}
\end{table}

\textbf{Which sensor carries the load information.} Table~\ref{tab:estimation} ablates head-mounted RGB-D, fingertip force, both, and both with arm kinematics. Under the force regression used here, force carries load information (0.757 versus 0.963\,kg for one nudge); RGB-D adds a small gain by observing interaction (0.729\,kg), and adding kinematics gives 0.660 versus 0.652\,kg for two nudges. Two orthogonal contacts reduce error to 0.652\,kg; we analyze second-contact geometry below.

\begin{table}[t]
\centering
\caption{Decision baselines on 3{,}618 test episodes. Accuracy uses the default protocol; the final columns use a cutoff selected at a 3\% validation risk target. For these columns, feature classifiers are also fitted on validation; test stays held out. All entries are percentages.}
\label{tab:decisions}
\footnotesize\setlength{\tabcolsep}{3pt}
\begin{tabular}{lrrr}
\toprule
Reader & Acc. & Test risk & Test missed \\
\midrule
\textsc{TacPhys}, tactile + RGB-D + kin. & 95.5 & 3.3 & 5.5 \\
\textsc{TacPhys}, tactile only & 95.3 & 3.2 & 6.1 \\
RGB-D only & 92.9 & 3.5 & 12.3 \\
Ideal resistance, \(\mu m g\) & 95.5 & 2.0 & 8.6 \\
\midrule
Peak force, first nudge & 85.3 & 2.5 & 32.3 \\
Mean peak force & 86.3 & 2.6 & 28.7 \\
Mean peak / true \(\mu\) & 86.7 & 3.2 & 27.2 \\
Logistic regression, features & 88.2 & 2.7 & 24.3 \\
Gradient boosting, features & 89.2 & 5.6 & 15.5 \\
\bottomrule
\end{tabular}
\end{table}

\textbf{Independent operating-point selection.} Table~\ref{tab:decisions} compares sequence inference with scalar thresholds and classifiers using log peaks, push speed, size, and contact offsets. Default learned decisions threshold predicted mass at 3\,kg; scalar and feature baselines use five-fold cross-validation. For the final columns, each cutoff is the largest of 2{,}000 validation-score quantiles meeting a 3\% validation risk target; reported risk and missed pushes are measured on test. The 4{,}162 validation episodes are session-disjoint from test. Feature classifiers are fitted on validation; learned-model cutoffs are selected per seed.

The full model misses 5.5\% of pushes at 3.3\% test risk, versus 28.7\% at 2.6\% for mean peak; even true friction leaves the peak baseline with 27.2\% misses at 3.2\% risk. A separate single-nudge control fits logistic regression, boosting, and an MLP on 29{,}202 training episodes and selects cutoffs on validation: test missed pushes are 28.8--31.5\% at 2.8--3.0\% risk, versus 7.7\% at 3.2\% risk for the five-seed tactile + RGB-D reader.

\textbf{Force--motion summary.} A least-squares fit \(F_n=c+\beta a_{\rm ee}\) over the first 0.6\,s of each nudge uses \(a_{\rm ee}=d\|v_{\rm ee}\|/dt\), an end-effector motion proxy. Thresholding the mean fitted intercept by five-fold cross-validation gives 91.7\% accuracy, improving on mean peak. At 3\% empirical test risk, it misses 16.5\% of pushes, versus 27.8\% for mean peak and 6.0\% for the full model. This diagnostic sweeps test scores, separately from Table~\ref{tab:decisions}'s validation-selected cutoffs.

\textbf{Architecture control.} Three tactile-only encoders share tokenization, conditioning, heads, and a 100-epoch schedule with early stopping over three seeds (Table~\ref{tab:encoders}). GRU and CNN use masked mean pooling instead of the transformer's \texttt{[CLS]} output. All reach approximately 95\% decision accuracy with fewer missed pushes than mean peak. Their MAEs lie 0.06--0.09\,kg above the 0.639\,kg resistance-only Bayes reference below, suggesting that mass information under unknown friction, rather than encoder capacity, chiefly limits accuracy. Their shared gain comes from integrating contact transients. We retain the transformer because its token interface accepts force, kinematics, RGB-D frames, and a second contact with a rotation embedding without architectural changes; GRU and CNN controls are tactile-only.

\begin{table}[t]
\centering
\caption{Controlled tactile-only encoders on the same 3{,}618 test episodes. MAE averages three seeds. Missed pushes use a diagnostic 3\% test-risk sweep. Parameters count only the temporal encoder. All three lie near the resistance-only reference in Sec.~\ref{sec:sim}.}
\label{tab:encoders}
\footnotesize\setlength{\tabcolsep}{3pt}
\begin{tabular}{lcccc}
\toprule
Encoder & Params & MAE (kg) & Acc. & Missed \\
\midrule
Transformer, 6 layers & 2.67M & \(0.694\pm0.007\) & 95.1\% & 7.3\% \\
BiGRU, 2 layers & 0.33M & \(0.724\pm0.014\) & 95.1\% & 7.3\% \\
1-D CNN, 4 blocks & 1.48M & \(0.733\pm0.013\) & 95.0\% & 8.3\% \\
\midrule
Mean-peak threshold & -- & -- & 86.3\% & 27.8\% \\
\bottomrule
\end{tabular}
\end{table}

\textbf{Contact onset and sensor compression.} Models trained with quasi-static speed ramps of 0.05 and 0.5\,s reach 94.1\% and 93.9\% accuracy, versus 88.7\% and 88.8\% for mean peak. A further two-seed control applies \(F\mapsto18\tanh(F/18)\) in newtons and per-nudge gain \(g\sim U(0.7,1)\) to both the model input and peak baseline. The full model is fine-tuned under this transformation. At 3\% empirical test risk, it misses 6.3--7.1\% of pushes versus 26.7\% for mean peak on 3{,}618 impulsive episodes; the corresponding rates on 626 quasi-static episodes are 7.1--12.2\% and 17.3\%. These sensor-control operating points are diagnostic test-score sweeps.

\textbf{Friction reference.} Quasi-static resistance scales with \(\mu_{\rm eff}mg\)~\citep{mason_pushing}. Under the simulated static-friction prior, a Bayes-optimal estimator observing only exact scalar resistance achieves 0.639\,kg MAE, close to the learned model's 0.652\,kg. The reference assumes log-uniform mass and the mean of two independent \(U[0.3,0.9]\) friction coefficients. It characterizes scalar-resistance inference; the model can also use contact transients and observed motion. Learned log-mass error retains a 0.35 partial correlation with friction after controlling for mass, motivating calibration under representative contact conditions.

\begin{figure}[t]
    \centering
    \includegraphics[width=\linewidth]{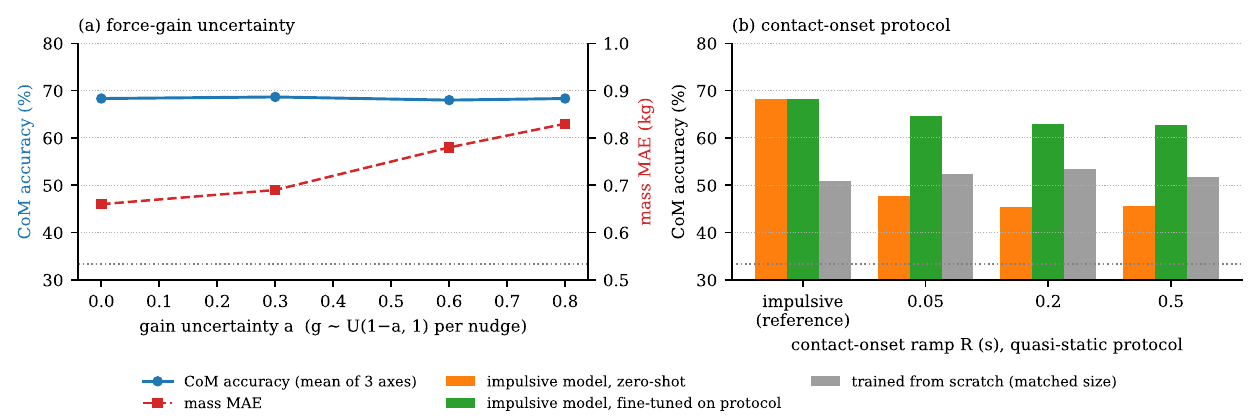}
    \caption{CoM sensitivity to force gain and contact onset (Sec.~\ref{sec:sim}). (a) CoM accuracy and mass error under per-nudge force-gain variation. (b) CoM transfer from impulsive contact to a creep--rest--ramp protocol, before and after fine-tuning.}
    \label{fig:comAB}
\end{figure}

\textbf{CoM identifiability and probe geometry.} On matched episodes, an orthogonal second contact reduces mass MAE from 0.719 to 0.644\,kg and raises lateral CoM accuracy from 69.8\% to 75.5\%; same-face repetition changes it from 71.6\% to 69.7\%. Two orthogonal contacts with tactile and visual inputs yield 71/74/66\% CoM accuracy. With per-nudge gains \(g\sim U(1-a,1)\) in training and testing, accuracy stays within one percentage point at \(a=0.8\), while mass MAE rises from 0.66 to 0.83\,kg. Transferring the impulsive model to quasi-static onset reduces CoM accuracy to 40/47/51\%; fine-tuning recovers 60--62/67--70/57--64\%. These results motivate an orthogonal second contact and identify impulsive onset as the hardware protocol change to test for CoM transfer. Figure~\ref{fig:comAB} summarizes the gain and onset comparisons.

\subsection{Hardware: Transfer, Site Calibration, and Online Decisions}
\label{sec:hardware}

\begin{figure*}[t]
    \centering
    \includegraphics[width=\linewidth]{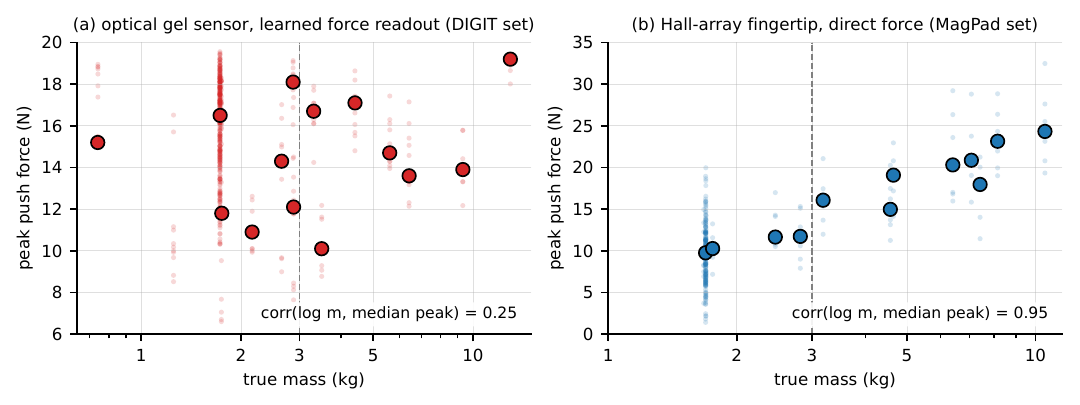}
    \caption{Hardware force readouts. Faint points are episodes; outlined points are object medians. The tested optical force regressor compresses peaks across loads. Hall-array readings track mass more closely. The dashed line marks 3\,kg.}
    \label{fig:sensors}
\end{figure*}

\textbf{Which force readout the decision needs.} A nudge must measure push resistance. With the tested DIGIT gel and learned force regressor, median peaks span 10--19\,N across 13 objects of 0.74--12.96\,kg (log-mass correlation 0.25); a leave-one-object-out force boundary achieves 17\% accuracy. The simulation-trained reader gives 2.6--3.2\,kg zero-shot MAE. An MLP residual with objects held out gives 1.6--1.9\,kg MAE and 51--53\% decision accuracy. Hall-array peaks correlate with log-mass at 0.95 (Fig.~\ref{fig:sensors}), supporting this front end for hardware evaluation.

\textbf{Sim-to-real transfer.} The simulation-trained tactile + RGB-D + kinematic \textsc{TacPhys} model reads hardware recordings without fine-tuning. Table~\ref{tab:hardware} reports zero-shot and affine MAE with each of 12 instances held out: up to 2.82\,kg, errors are 0.28--1.07\,kg zero-shot and 0.43--0.61\,kg after calibration; the 3.19\,kg waste bin gives 1.27 and 0.69\,kg; above 4\,kg, errors are 1.9--7.2\,kg and 1.3--6.5\,kg. On 220 orthogonal pairs with containers held out, zero-shot and affine decision accuracies are 88.2\% and 88.6\%.

\begin{table}[t]
\centering
\caption{Hall-array results with each mass instance held out. ZS and Aff.\ are zero-shot and affine-calibrated \textsc{TacPhys} MAE (kg) over the 220 orthogonal pairs. Rise is the mean force increment (N); Acc.\ is the calibrated-boundary accuracy over all 262 pairs, and Agree gives correct / accepted episodes under offline agreement. Same-letter boxes share a container.}
\label{tab:hardware}
\footnotesize\setlength{\tabcolsep}{2.5pt}
\begin{tabular}{lrrrrrr}
\toprule
Instance & kg & ZS & Aff. & Rise & Acc. & Agree \\
\midrule
Cube & 1.69 & 0.45 & 0.43 & 5.3 & 96\% & 163/165 \\
Box E1 & 1.76 & 0.28 & 0.49 & 8.1 & 83\% & 2/2 \\
Box A1 & 2.46 & 0.71 & 0.57 & 6.9 & 100\% & 4/4 \\
Box D1 & 2.82 & 1.07 & 0.61 & 8.4 & 100\% & 3/3 \\
Waste bin & 3.19 & 1.27 & 0.69 & 13.5 & 67\% & 4/4 \\
Box D2 & 4.58 & 2.54 & 1.96 & 12.8 & 50\% & 1/1 \\
Box B1 & 4.65 & 1.92 & 1.25 & 16.1 & 100\% & 5/5 \\
Box B2 & 6.41 & 3.65 & 2.97 & 18.0 & 100\% & 5/5 \\
Box A2 & 7.08 & 4.72 & 4.17 & 19.5 & 100\% & 6/6 \\
Box E2 & 7.42 & 4.88 & 4.23 & 12.4 & 40\% & 1/2 \\
Box C1 & 8.15 & 4.41 & 3.57 & 19.8 & 100\% & 5/5 \\
Box C2 & 10.52 & 7.23 & 6.45 & 22.3 & 100\% & 5/5 \\
\midrule
All (262 pairs) & & & & & 93.5\% & 204/207 \\
\bottomrule
\end{tabular}
\end{table}

\textbf{Site calibration.} The boundary \(\tau_F\) (Sec.~\ref{sec:calibration}) maximizes calibration accuracy over 4--30\,N at 0.25\,N spacing with at most 30 cube episodes per fold; instance-fold values span 10.9--12.3\,N. Accuracy is 93.5\% on all 262 instance-held-out pairs (Table~\ref{tab:hardware}) and 92.7\% on 220 orthogonal container-held-out pairs. Non-cube accuracy is 86.4\% (always-detour: 71.2\%); always-push achieves 82.1\% overall (215/262). These force-rise results are offline. The online trials used a separately calibrated raw-peak rule fixed before this analysis.

Agreement accepts 207/262 episodes with 204 correct (98.6\% accuracy, 79.0\% coverage). It authorizes 1/33 accepted heavy episodes and detours 2/174 accepted light episodes, deferring 14 heavy and 41 light episodes. Without agreement, heavy errors are 8/47 and light detours 9/215. Counting deferral as no push raises light-object non-authorization from 9/215 (4.2\%) to 43/215 (20.0\%); non-cube coverage is 42/66 (63.6\%). Without and with agreement, heavy-error Wilson 95\% intervals are 8.9--30.1\% and 0.5--15.3\%. A 5{,}000-draw bootstrap over seven containers with frozen instance-fold predictions gives 76.8--96.5\% accuracy without agreement and 93.5--100\% with it; heavy-error intervals are 3.3--41.4\% and 0--14.3\%.

\textbf{Floor and friction.} The reading scales with \(\mu_{\rm eff}mg\). Carpet and plastic shift light-container implied friction by 25--30\% (cube 0.61 versus 0.47, box A1 0.67 versus 0.51); carpet-to-plastic force-rise ratios span 0.71--1.27. A boundary fitted on one floor decides 93.2\% on the other (93.1\% in reverse). Heavy authorizations occur only for three objects near the boundary or with non-rigid payloads (waste bin 2/6, D2 3/6, E2 3/5). Other heavy instances achieve 100\% on both floors; the cube achieves 95\% on carpet and 100\% on plastic. At \(\tau=2\)\,kg, restricting orthogonal pairs to masses at most 3.2\,kg leaves 187 episodes. Lighter E1 produces a larger force rise than A1 and D1; refitted-boundary accuracy is 42.2\%, and zero-shot accuracy is 83.4\%, but 12/18 heavy episodes are authorized. This overlap motivates calibration around the main 3\,kg boundary and is consistent with simulation's 0.35 partial correlation between log-mass error and friction.

\textbf{Online decision evaluation.} The standing robot probed two visually identical unseen boxes of 1.47 and 5.30\,kg, issuing push-or-detour judgments over five paired episodes per box and floor. The force boundary was calibrated on collection objects and fixed before testing (14\,N on mean peak), deciding 15/20 episodes correctly. Errors were two heavy authorizations on plastic and three light rejections on carpet, all within 1.5\,N of the boundary; the floor shifts readings by about 30\%. Exploratory per-floor fits give 19/20 on these same trials. Offline, \textsc{TacPhys} estimates 0.8--1.9\,kg for the light box and 1.3--2.4\,kg for the heavy box: light decisions are correct; heavy under-reading reflects the sensor-range effect discussed in Sec.~\ref{sec:limitations}.

\subsection{Value in Repeated Patrol}
\label{sec:patrol}

\begin{figure*}[t]
    \centering
    \includegraphics[width=\linewidth]{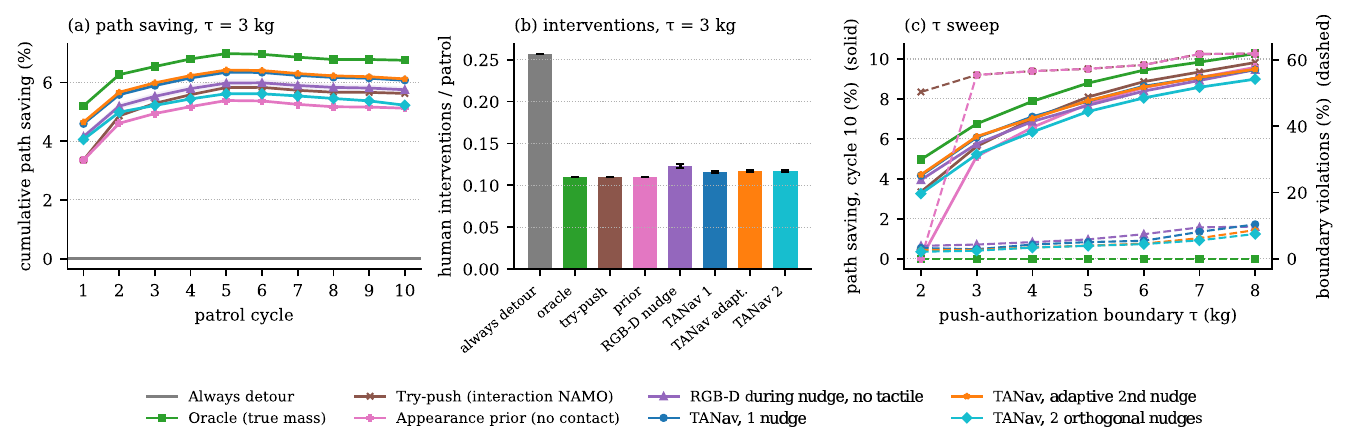}
    \caption{Repeated-patrol simulation. (a) Cumulative path saving at 3\,kg. (b) Human interventions per patrol. (c) Path saving (solid) and boundary violations (dashed) as the authorization boundary varies. All policies share the planner; cached estimates are reused across cycles.}
    \label{fig:patrol}
\end{figure*}

\textbf{Protocol.} We evaluate 300 randomized layouts with 6\(\pm\)2 route blockers and 10 patrol cycles. Results are summarized in Table~\ref{tab:patrol} and Fig.~\ref{fig:patrol}; an example layout is shown in Fig.~\ref{fig:layout}.

\begin{table}[t]
\centering
\caption{Repeated patrol over 300 layouts and ten cycles, with five seeds and a common planner at 3\,kg. Saving is at cycle ten relative to always-detour; interventions are per patrol and failures per scene. Agreement variants are defined in Sec.~\ref{sec:calibration}.}
\label{tab:patrol}
\footnotesize\setlength{\tabcolsep}{3pt}
\begin{tabular}{lcccc}
\toprule
Policy & Saving & Interv. & Risk & Failed \\
\midrule
Always detour & 0\% & 0.26 & 0\% & 0 \\
Oracle (true mass) & 6.8\% & 0.11 & 0\% & 0 \\
Try-push & 5.6\% & 0.11 & 55.5\% & 2.6 \\
Appearance prior & 5.1\% & 0.11 & 55.5\% & 5.3 \\
RGB-D during nudge & 5.8\% & 0.12 & 4.3\% & 0.39 \\
\textsc{TANav}, one nudge & 6.1\% & 0.12 & 2.9\% & 0.33 \\
\textsc{TANav}, adaptive second & 6.1\% & 0.12 & 2.7\% & 0.26 \\
\textsc{TANav}, two nudges & 5.2\% & 0.12 & 2.4\% & 0.23 \\
\textsc{TANav}, sequential agree. & 5.1\% & 0.12 & 1.6\% & 0.17 \\
\textsc{TANav}, contact agreement & 5.0\% & 0.12 & \textbf{1.0\%} & \textbf{0.11} \\
\midrule
Peak, one nudge & 5.4\% & 0.13 & 10.3\% & 0.96 \\
Peak, two nudges & 4.7\% & 0.13 & 6.7\% & 0.61 \\
Peak, contact agreement & 4.2\% & 0.14 & 4.4\% & 0.42 \\
\bottomrule
\end{tabular}
\end{table}

Blockers draw masses, geometry, and held-out predictions from the common simulation episode pool; all policies see the same object sequences. Between cycles, an object is replaced with probability 0.3. At 25\% of chokepoints no detour is available, so a non-push decision requests human assistance with a 300\,s wait. Navigation speed is 0.8\,m/s and pushing speed 0.3\,m/s. Probe costs include 5\,s alignment, 1.42\,s contact, and 1\,s retraction, plus 4\,s realignment for a second contact. A route-clearing push adds 3\,s setup and 2\,m extra travel; a failed attempt adds 3\,s.

Patrol uses tactile + RGB-D mass estimates at the fixed boundary, rather than the validation-selected cutoffs of Table~\ref{tab:decisions}. Peak policies use log-linear mass maps fitted on validation predictions and frozen for test. Violations count authorizations at the first decision for each encountered heavy object instance, failed pushes are per scene across ten cycles, and missed pushes count feasible, eligible actions whose true cost is below detouring. Try-push remembers failures; the appearance prior can retry the same object, giving equal risk but different failure counts. Independent mass--appearance sampling removes load-specific cues from this prior.

\textbf{Navigation benefit.} One-nudge \textsc{TANav} recovers 90\% of the oracle's path saving (6.1\% versus 6.8\%) and raises the fraction of autonomous patrol cycles, defined as cycles completed without human intervention, from 78.1\% to 89.3\% relative to always-detour, with interventions falling from 0.26 to 0.12 per patrol (Table~\ref{tab:patrol}). Tactile input primarily reduces risk relative to RGB-D-only probing: path savings are similar, while violations fall from 4.3\% to 2.9\% and failed pushes from 0.39 to 0.33 per scene. The paired path-saving difference is 0.32 percentage points (95\% bootstrap interval over 300 layouts: [0.16, 0.50]); cumulative time over ten cycles, including probes and intervention waits, is 1569 versus 1595\,s. Try-push reaches 0.11 interventions but authorizes 55.5\% of heavy objects. The one-nudge peak policy gives 5.4\% saving and 10.3\% violations, with 0.96 failed pushes per scene.

\textbf{Additional contact.} Adaptive second contacts target estimates near the boundary: they retain 6.1\% path saving while reducing failed pushes from 0.33 to 0.26 per scene, using 6.6 probes versus 10.2 for fixed two-contact probing. Two fixed nudges reduce violations to 2.4\%. Per-contact agreement further reduces them to \(1.0\pm0.2\%\), with \(0.11\pm0.03\) failed pushes per scene, at a cost of 0.2 percentage points in path saving. Peak-force agreement yields 4.4\% violations and 0.42 failed pushes. Learned agreement also misses fewer useful pushes, 27.2\% versus 37.2\%.

The \(\tau=2\)--8\,kg sweep (Fig.~\ref{fig:patrol}c) shows the resulting trade-off: one-nudge \textsc{TANav} stays within one percentage point of oracle path saving, with higher violation rates at larger boundaries. A direct classifier with the same encoder, trained to predict \(m>3\,\mathrm{kg}\) and feeding its probability to the cost rule, achieves similar performance there (5.9\% path saving, 2.0\% violations), but applying it unchanged at 2\,kg gives 10.3\% violations versus 3.2\% for regression. At 4--8\,kg its saving stays near 6\%, versus 7--9\% for regression; this tests reuse of a fixed-boundary classifier, not retraining for each boundary.

\textbf{Observation reuse.} Re-probing every encounter leaves mean path saving and violations unchanged at the reported precision, but increases probes from 5.86 to 8.50 per scene and reduces time saving by 1.04 percentage points. In a separate 300-layout sensitivity set, missing 25/50/100\% of replacements raises violations from 3.3\% to 5.8/9.0/16.9\%; re-probing stays at 3.3\%, motivating reassociation checks or cache invalidation.

\section{Discussion}
\label{sec:discussion}

In simulation, learned mass supports both push authorization and patrol cost. Site calibration relates resistance to the handling boundary under local contact conditions; mass and friction remain coupled. A change in \(\tau\) therefore calls for calibration around the new boundary, as the 2\,kg evaluation shows.

\section{Limitations}
\label{sec:limitations}

Hardware calibration covers seven containers on two floors, with most episodes collected on the cube. Beyond the fingertip's range, mass estimates compress: with containers held out, the affine reader authorizes 20/39 heavy episodes versus 8/39 for the calibrated boundary. A wider-range fingertip is needed to extend mass estimation to the site's heavier loads. Real-cube CoM accuracy is 25--40\% per axis under quasi-static probing; impulsive onset is the protocol change to test. Further hardware evaluation will address autonomous contact placement, learned-reader decisions online, and object reassociation across patrol visits.

\section{Conclusion}

\textsc{TANav} connects a diagnostic nudge to navigation through push resistance, mass estimation, site-defined authorization, and route cost. In simulation, \textsc{TacPhys} approaches a resistance-only Bayes reference and reduces missed pushes from 28.7\% to 5.5\%; repeated patrol recovers 90\% of the oracle's path saving and raises autonomous cycles from 78.1\% to 89.3\%. Offline hardware tests show zero-shot transfer on light containers and 93.5\% accuracy with force-rise calibration; a separate raw-peak rule gives 15/20 correct online judgments on unseen boxes.

\clearpage
\bibliography{example}

\clearpage
\appendix
\raggedbottom

\section{Supplementary Views}

\begin{figure}[ht]
    \centering
    \includegraphics[width=0.78\linewidth]{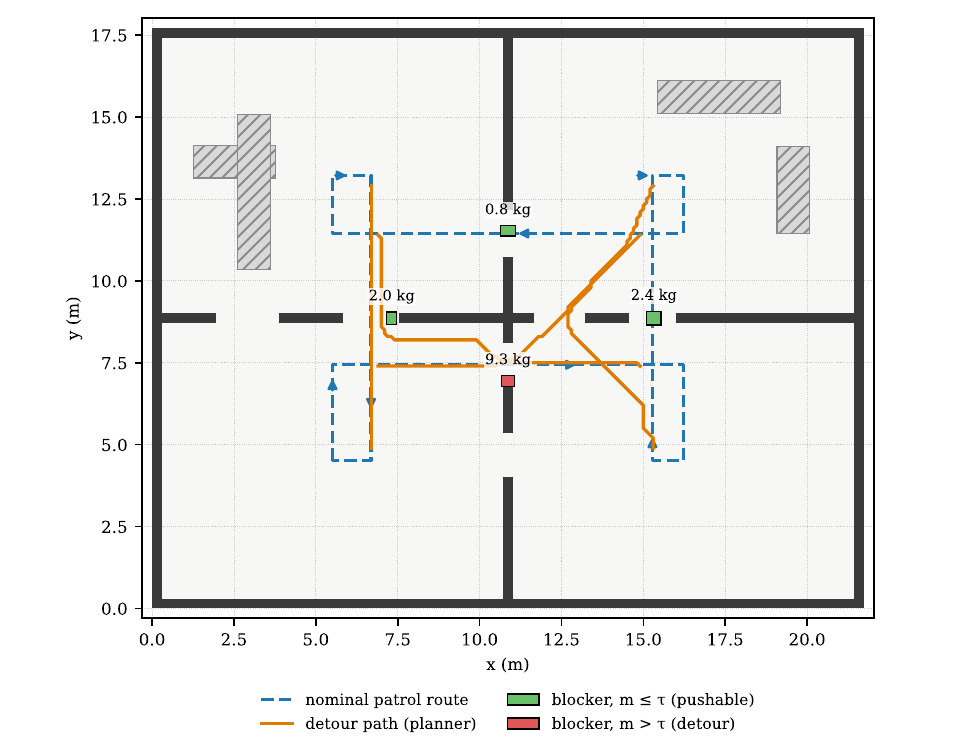}
    \caption{Example layout from the repeated-patrol simulation (Sec.~\ref{sec:patrol}). The nominal route passes through doorways containing blockers; orange paths show detours. Green and red markers distinguish loads at or below the 3\,kg boundary from heavier loads.}
    \label{fig:layout}
\end{figure}

\FloatBarrier
\end{document}